\documentclass[a4paper]{article}
\usepackage{amsmath,epsfig}

\usepackage{epsf}
\usepackage{times}
\usepackage{float}
\usepackage{graphicx}
\usepackage{subcaption}
\usepackage{tabularx}

\usepackage{amssymb}
\usepackage{amsmath}
\usepackage{subeqnarray}
\usepackage{color}

\usepackage{enumitem}

\twocolumn
\def\ninept{\def\baselinestretch{.95}\let\normalsize\small\normalsize}
\title{Deep Speaker: an End-to-End Neural Speaker Embedding System}

\author{Chao Li$^*$, Xiaokong Ma$^*$, Bing Jiang$^*$, Xiangang Li \thanks{equally contributed to this work} \\
Xuewei Zhang, Xiao Liu, Ying Cao, Ajay Kannan, Zhenyao Zhu}
\date{
Baidu Inc. %\hskip 10pt \\
}

\begin{document}
\ninept
\begin{sloppy}
\maketitle
%\vskip -10pt
\begin{abstract}
We present Deep Speaker, a neural speaker embedding system that maps utterances to a hypersphere where speaker similarity is measured by cosine similarity.
The embeddings generated by Deep Speaker can be used for many tasks, including speaker identification, verification, and clustering.
We experiment with ResCNN and GRU architectures to extract the acoustic features, then mean pool to produce utterance-level speaker embeddings, and train using triplet loss based on cosine similarity.
Experiments on three distinct datasets suggest that Deep Speaker outperforms a DNN-based i-vector baseline. For example, Deep Speaker reduces the verification equal error rate by 50\% (relatively) and improves the identification accuracy by 60\% (relatively) on a text-independent dataset.
We also present results that suggest adapting from a model trained with Mandarin can improve accuracy for English speaker recognition.
\end{abstract}

%\vspace{10pt}
\section{Introduction}
\label{sec:intro}

Speaker recognition algorithms seek to determine the identity of a speaker from audio.
Two common recognition tasks are speaker verification (determining whether a speaker's claimed identity is true or false) and speaker identification (classifying the identity of an unknown voice among a set of speakers).
Verification and identification algorithms may require the speaker to utter a specific phrase (text-dependent recognition) or be agnostic to the audio transcript (text-independent recognition). 
In all these subtasks, embedding methods can be used to map utterances into a feature space where distances correspond to speaker similarity.
Though many algorithms have pushed the state-of-the-art over the past couple years \cite{svm}\cite{jfa}\cite{ivector}\cite{dnn-ivector}\cite{plda}\cite{okgoogle}\cite{povy}, speaker
 recognition is still a challenging task.

The traditional speaker recognition approach entails using i-vectors \cite{ivector} and probabilistic linear discriminant analysis (PLDA) \cite{plda}.
This framework can be decomposed into three stages \cite{dnn-ivector}:
\setlist{before=\normalfont,font=\itshape} 
\vspace{-3pt}
\begin{description}
   \item[~~~~Step 1] Collect sufficient statistics
  \vspace{-3pt}
   \item[~~~~Step 2] Extract speaker embeddings (i-vector)
  \vspace{-3pt}
   \item[~~~~Step 3] Classify (PLDA)
\end{description}
\vspace{-3pt}
Sufficient statistics (also known as Baum-Welch statistics) are computed from a Gaussian Mixture Model-Universal Background Model (GMM-UBM), which is optimized using a sequence of feature vectors (e.g., mel-frequency cepstral coefficients (MFCC) \cite{ivector}.
Recently, deep neural network (DNN) acoustic models have also been used to extract sufficient statistics \cite{dnn-ivector}.
The high-dimensional statistics are converted into a single low-dimensional i-vector that encodes speaker identity and other utterance-level variability.
A PLDA model is then used to produce verification scores by comparing i-vectors from different utterances \cite{plda}.

The three steps of an i-vector system are traditionally trained on subtasks independently, not jointly optimized.
An alternative DNN-based approach uses a classification layer \cite{softmax}, combining both \textit{Step 1} and \textit{Step 2}.
The intermediate bottleneck layer in the DNN provides a frame-level embedding, which can be used for speakers not included in the training set.
During prediction, additional steps are required to aggregate frame-level representations and to perform verification.
This approach suffers from at least two major issues: (1) \textit{Step 1} and \textit{Step 2} are not directly optimized with respect to speaker recognition, and (2) there's a mismatch between training and test. The training labels are given at the frame-level, while utterance-level predictions are made in testing.

\cite{okgoogle} and \cite{povy} introduced end-to-end neural speaker verification systems, combining all three steps.
\cite{okgoogle} used the last frame output of a long short-term memory (LSTM) \cite{lstm-proposed} model as an utterance-level speaker embedding,
while \cite{povy} used a network-in-network (NIN) \cite{nin} nonlinearity followed
by an utterance-level pooling layer to aggregate frame-level representations. Both \cite{okgoogle}
and \cite{povy} were trained using the same distance metric.

In this paper, we extend the end-to-end speaker embedding systems proposed in \cite{okgoogle} and \cite{povy}.
First, a deep neural network is used to extract frame-level features from utterances.
Then, pooling and length normalization layers generate utterance-level speaker embeddings.
The model is trained using triplet loss \cite{facenet}, which minimizes the distance between embedding pairs from the same speaker and maximizes the distance between pairs from different speakers. Pretraining using a softmax layer and cross-entropy over a fixed list of speakers improves model performance.

More specifically, we test convolutional neural network (CNN)-based and recurrent neural network (RNN)-based architectures for frame-level feature extraction, and present results both for speaker verification and speaker identification.
CNNs are effective for reducing spectral variations and modeling spectral correlations in acoustic features \cite{deepcnn-hmm}.
CNNs have also recently been applied to speech recognition with good results  \cite{deepcnn-hmm}\cite{lace}\cite{vgg}\cite{deepcnn-ctc}.
Since deep networks can better represent long utterances than shallow networks \cite{deepcnn-ctc}, we propose a deep residual CNN (ResCNN), inspired by residual networks (ResNets) \cite{resnet}.  We also investigate stacked gated recurrent unit (GRU) \cite{gru-proposed} layers as an alternative for frame-level feature extraction, since they have proven to be effective for speech processing applications \cite{ds2}\cite{cldnn}.

Like \cite{povy}, we use a distance-based loss function to discriminate between same-speaker and different-speaker utterance pairs.
However, unlike the PLDA-like loss function in \cite{povy}, we train our networks so that cosine similarity in the embedding space directly corresponds to utterance similarity.
We also select hard negative examples at each iteration by checking candidate utterances globally, not just in the same minibatch.
This approach provides faster training convergence.

 \begin{figure*}[t]
  \centering
  \centerline{\includegraphics[width=17.5cm]{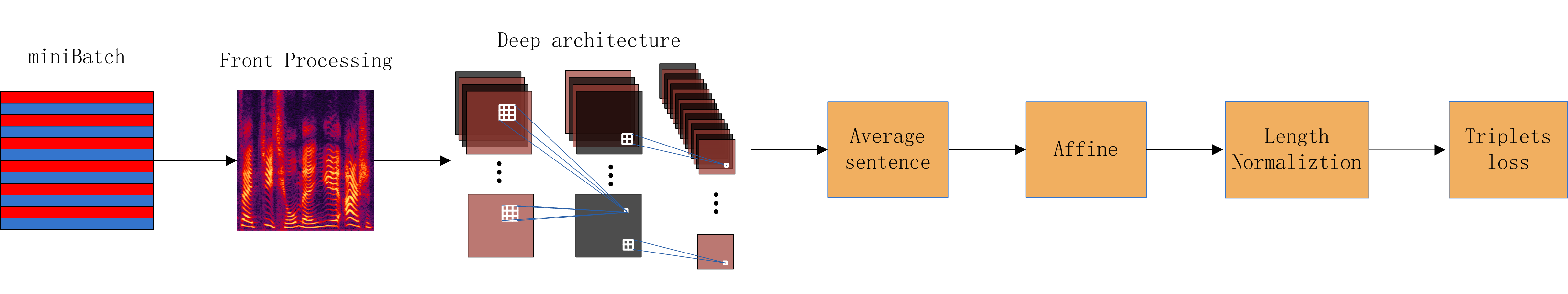}}
\caption{{\it Diagram of the Deep Speaker architecture}}
\label{fig:archit}
\end{figure*}

Finally, we evaluate our proposed Deep Speaker system on three different datasets, for text-independent and text-dependent speaker recognition tasks in both Mandarin and English. We also investigate the effects of softmax pre-training, system combination, training dataset size, and enrollment utterance count. The experiments indicate Deep Speaker can significantly improve over DNN-based i-vector text-independent speaker recognition systems. In the text-dependent task, Deep Speaker can match a DNN i-vector baseline system, and improve upon it if fine-tuned after text-independent training.
In particular, two interesting results are shown: (1) Deep Speaker leverages big data well (performance boosts when trained on as many as 250,000 speakers), and (2) Deep Speaker can transfer well across spoken languages that are vastly different, \textit{i.e.}, Mandarin and English.

%\vspace{10pt}

 %\vspace{10pt}
\section{Related Work}
\label{sec:related work}
Traditionally, i-vectors have been used to model inter-speaker variability \cite{ivector}.
i-vector-based speaker recognition models perform classification using cosine similarity between i-vectors or more advanced techniques such as PLDA \cite{plda-ori},
heavy-tailed PLDA \cite{plda-ht}, and Gauss-PLDA \cite{plda}.

There have been several papers replacing pieces of the traditional speaker recognition system with DNNs.
One approach is to train a GMM on bottleneck features extracted from a DNN, and then extract i-vectors \cite{bn}.
Another DNN-based approach uses an acoustic speech recognition DNN instead of a UBM-GMM to produce frame posteriors for i-vector computation \cite{dnn-ivector}.
Ehsan Variani \emph{et al.} \cite{softmax} trained DNNs to classify speakers with frame-level acoustic features.
The activations of the final hidden layer are averaged over the utterance to create a ``d-vector'' which replaces the i-vector.
These approaches all show improvements upon the traditional i-vector baseline.

There have recently been end-to-end neural speaker recognition efforts as well.
Georg Heigold \emph{et al.} \cite{okgoogle} trained an LSTM for text-dependent speaker verification, which acheived a
$2\%$ equal error rate (EER) on the ``Ok Google'' benchmark.
The model maps a test utterance and a few reference utterances directly to a single score for verification and jointly optimizes the system's components using the same evaluation protocol as at test time.
David Snyder \emph{et al.} \cite{povy} also train an end-to-end text-independent speaker verification system.
Like \cite{okgoogle}, the objective function separates same-speaker and different-speaker pairs, the same scoring done during verification. The model reduces EER by $13\%$ on average, compared to the i-vector baseline.

Our paper uses different architectures than \cite{okgoogle} and \cite{povy} that balance inference time with model depth and also draw from state-of-the-art speech recognition systems.
We showcase our models' efficacy on both text-dependent and text-independent speaker recognition tasks.
Lastly, we provide novel insight on the effect of dataset size, softmax pre-training, model fusion, and adaptation from one language to another.

\section{Deep Speaker}
\label{sec:proposed method}

Figure \ref{fig:archit} illustrates the architecture of Deep Speaker. Raw audio is first preprocessed using the steps detailed in Section \ref{sec:training}.
Then, we use a feed-forward DNN to extract features over the preprocessed audio. We experiment with
two different core architectures: a ResNet-style \cite{resnet} deep CNN and the Deep Speech 2 (DS2)-style
architecture consisting of convolutional layers followed by GRU layers.
The details of these networks are described in Section \ref{sec:dnn}.
A sentence average layer converts frame-level input to an utterance-level speaker representation.
Then, an affine layer and a length normalization layer map the temporally-pooled features to a speaker embedding, as presented in Section \ref{sec:embedding}.
Finally, the triplet loss layer operates on pairs of embeddings, by maximizing the cosine similarities of embedding pairs from the same speaker, and minimizing those from different speakers, as
explained in Section \ref{sec:triplet loss}.

\subsection{Neural Network Architecture}
\label{sec:dnn}

As stated above, we use two types of deep architectures for frame-level audio feature extraction.

\subsubsection{Residual CNN}
\label{sec:resnet}

\begin{figure}[t]
\centering
\includegraphics[width=2cm]{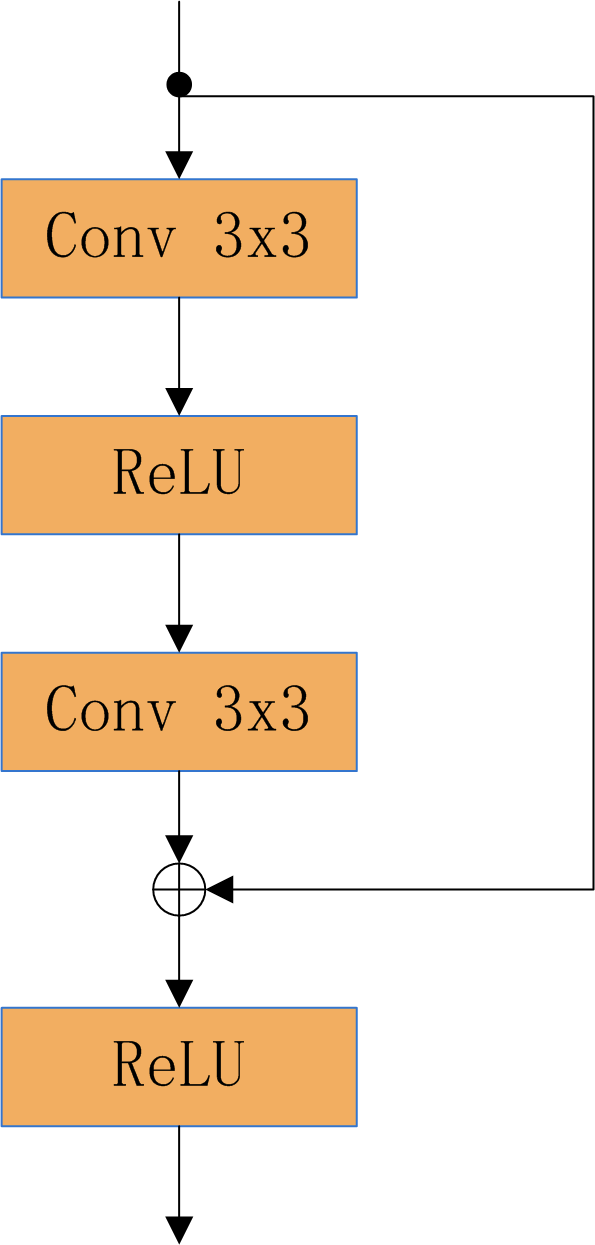}
%\vspace{-20pt}
\caption{\it Detailed view of ResBlock. A convolution block Conv $3\times3$ is parameterized by the filter size  $3\times3$, the zero padding 1 in both directions and the consecutive striding  $1\times1$}
\label{fig:resnet block}
\end{figure}

Though deep networks have larger capacity than shallow networks, they tend to be more difficult to train.
ResNet \cite{resnet} was proposed to ease the training of very deep CNNs.
ResNet is composed of a number of stacked residual blocks (ResBlocks).
Each ResBlock contains direct links between the lower layer outputs and the higher layer inputs, as described in Figure \ref{fig:resnet block}.
The ResBlock is defined as
\begin{equation}
\label{equ:resnet block}
h = \digamma(x, W_i)+ x,
\end{equation}
where $x$ and $h$ are the input and output of the layers considered, and $\digamma$ is the stacked nonlinear layer's mapping function.
Note that identity shortcut connections of $x$ do not add extra parameters and computational complexity.

Table \ref{tab:resnet struct} shows the details of the proposed ResCNN architecture.
As described in Figure \ref{fig:resnet block}, the ResBlock contains two convolutional layers with $3\times3$ filters and $1\times1$ stride.
Each block has an identical structure, and the skip connection is the identity mapping of $x$.
Three ResBlocks are stacked in our architecture.
When the number of channels increases, we use a single convolutional layer with filter size of $5\times5$ and a stride of $2\times2$.
Thus, the the frequency dimension is kept constant in all convolution layers.
We find that speaker recognition performance is not sensitive to the stride in the time dimension, contrary to \cite{deepcnn-ctc}'s findings.
Notably, when the number of channels increases, projection shortcuts are not used as in \cite{resnet}, because they increased the number of parameters without yielding significant improvement.
We adopt sequence-wise batch normalization (BN) between the convolution and the nonlinearity, following \cite{ds2}.
We use the clipped rectified linear (ReLU) function \cite{ds2},

\begin{equation}
\label{equ:relu}
\sigma (x) = \min \left\{ {\max \left\{ {x,0} \right\},20} \right\}
\end{equation}
as our nonlinearity for all of the network layers.

\begin{table}[t]
\caption{\it Architecture of ResCNN. ``Average'' denotes the temporal pooling layer and ``ln'' denotes the length normalization layer. A bracket describes the structure of a ResBlock as shown in  Fig. \ref{fig:resnet block}}
\footnotesize
\label{tab:resnet struct}
%\vspace{-2mm}
\begin{center}
\begin{tabular}{c|c|c|c|c}
\hline

layer name & structure & stride & dim & \# params \\
\hline %\vspace{1pt}
\hline

conv64-s & $5\times5,64$ & 2$\times2$ & 2048 & 6K \\
%\hline  \vspace{1pt}
res64 & $\left[ {\begin{array}{*{20}{c}}
{3 \times 3,64}\\
{3 \times 3,64}
\end{array}} \right] \times 3 $ & 1$\times1$ & 2048 & 41K$\times6$ \\
\hline

conv128-s & $5\times5,128$ & 2$\times2$ & 2048 &  209K\\
%\hline \vspace{1pt}
res128 & $\left[ {\begin{array}{*{20}{c}}
{3 \times 3,128}\\
{3 \times 3,128}
\end{array}} \right] \times 3$ & 1$\times1$ & 2048 & 151K$\times6$ \\
\hline

conv256-s & $5\times5,256$ &2$\times2$ & 2048 & 823K \\
%\hline \vspace{1pt}
res256 & $\left[ {\begin{array}{*{20}{c}}
{3 \times 3,256}\\
{3 \times 3,256}
\end{array}} \right] \times 3$ &1$\times1$ & 2048 & 594K$\times6$ \\
\hline

conv512-s & $5\times5,512$ &2$\times2$ & 2048 & 3.3M \\
%\hline \vspace{1pt}
res512 & $\left[ {\begin{array}{*{20}{c}}
{3 \times 3,512}\\
{3 \times 3,512}
\end{array}} \right]\times 3$ & 1$\times1$ & 2048 & 2.4M$\times6$ \\
\hline

average & - & - & 2048 & 0 \\
\hline
affine & $2048\times512$ & - & 512 & 1M \\
\hline
ln &  - & - & 512 & 0 \\
\hline
triplet &  - & - & 512 & 0 \\
\hline
\hline
total &   &  &  & 24M \\
\hline
\end{tabular}%
\end{center}
\end{table}

\begin{table}[t]
\caption{\it Architecture of the GRU model.  ``Average'' denotes the temporal pooling layer, and ``ln'' denotes the length normalization
layer.}
\footnotesize
\label{tab:gru struct}
%\vspace{-2mm}
\begin{center}
\begin{tabular}{c|c|c|c|c}
\hline

layer name & struct & stride & dim & param \\
\hline %\vspace{1pt}
\hline

conv64-s & $5\times5,64$ & 2$\times2$ & 2048 & 6K \\
\hline  %\vspace{1pt}

GRU & 1024 cells  & 1 & 1024 & 9.4M \\
\hline  %\vspace{1pt}

GRU &1024 cells  & 1 & 1024 & 6.3M \\
\hline  %\vspace{1pt}

GRU & 1024 cells  & 1 & 1024 & 6.3M \\
\hline  %\vspace{1pt}

average & - & - & 1024 & 0 \\
\hline
affine & $1024\times512$ & - & 512 & 500K \\
\hline
ln &  - & - & 512 & 0 \\
\hline
triplet &  - & - & 512 & 0 \\
\hline
\hline
total &   &  &  & 23M \\
\hline
\end{tabular}%
\end{center}
\end{table}

\subsubsection{GRU Network}
\label{sec:bigru}

We also experiment with recurrent networks for frame-level feature extraction because they have worked well for speech recognition \cite{cldnn}.
\cite{gru vs lstm} showed that a GRU is comparable to an LSTM with a properly initialized forget gate bias, and their best variants are competitive with each other.
We decided to use GRUs because previous speech recognition experiments \cite{ds2} on smaller data sets showed the GRU and LSTM reach similar accuracy for the same number of parameters, but the GRUs were faster to train and less likely to diverge.

The details of the proposed GRU architecture are shown in Table \ref{tab:gru struct}. A $5\times5$ filter size, $2\times2$ stride convolution layer (like in the ResCNN architecture), reduces dimensionality in both the time and
frequency domains, allowing for faster GRU layer computation.
Following the convolutional layer are three forward-only GRU layers with 1024 units, recurrent in the time dimension.
After the GRU layers, we apply the same average, affine, and length normalization layers as in the ResCNN model. We also use sequence-wise batch normalization and clipped-ReLU
activation in the whole model.

The ResCNN and GRU architectures have a similar number of parameters, 23M-24M, allowing us to better compare their performances.

\subsection{Speaker Embedding}
\label{sec:embedding}

Frame-level activations are fed into a temporal average layer.
Unlike in the pooling layer in \cite{povy}, we do not use standard deviation of frame-level outputs.
The layer activation \textit{h} is computed as follows:

\begin{equation}
\label{equ:average}
h = \frac{1}{T}\sum\limits_{t = 0}^{T - 1} {x(t)}
\end{equation}
where, $T$ is the number of frames in the utterance.
An affine layer then projects the
utterance-level representation into a 512-dimensional embedding.
We normalize embeddings to have unit norm and use cosine similarity between pairs in the objective function:

\begin{equation}
\label{equ:distance}
\cos ({x_i},{x_j}) = x_i^{\rm{T}}{x_j}
\end{equation}
where, $x_i$ and $x_j$ are two embeddings.

\subsection{Triplet Loss and Selection}
\label{sec:triplet loss}

We model the probability of embeddings $x_i$ and $x_j$ belonging to the same speaker by their cosine similarity in Equation (\ref{equ:distance}), allowing us to use the triplet loss function like
in FaceNet \cite{facenet}.

\begin{figure}[t]
\centering
\includegraphics[width=8.5cm]{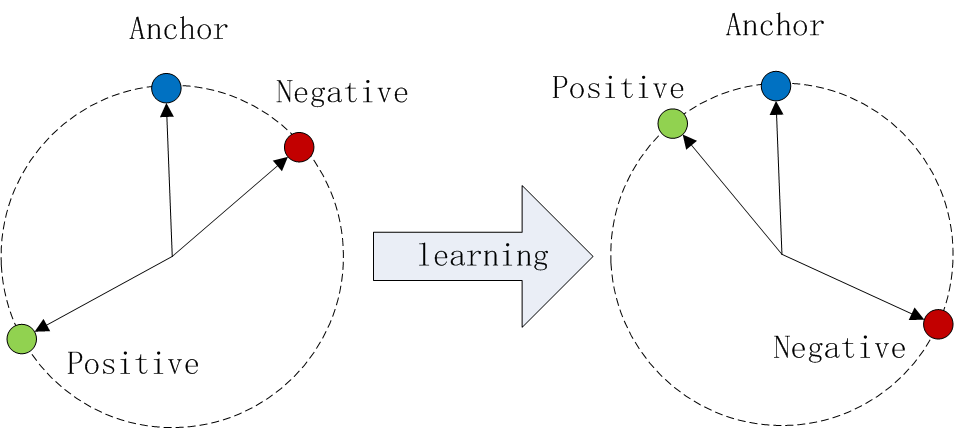}
%\vspace{-20pt}
\caption{\it The Triplet loss in cosine similarity.}
\label{fig:triplet}
\end{figure}

As shown in Figure \ref{fig:triplet}, triplet loss takes in as input three samples, an anchor (an utterance from a specific speaker), a positive example (another utterance from the same speaker), and a negative example (an utterance from another speaker).  We seek to make updates such that the cosine similarity between the anchor and the positive example is larger than the cosine similarity between the anchor and the negative example \cite{facenet}.  Formally,

\begin{equation}
  \label{equ:st.}
{s_i^{ap}} - \alpha > {s_i^{an}}
\end{equation}
where, $s_i^{ap}$ is the cosine similarity between the anchor $a$ and the positive example $p$ for triplet $i$, and $s_i^{an}$ is the cosine similarity between the anchor $a$ and the negative example $n$ in triplet $i$. We impose a minimum margin $\alpha$ between those similarities.  The cost function for $N$ triplets can be written as

  \begin{equation}
  \label{equ:cost}
L = \sum\limits_{i = 0}^N {{{\left[ {s_i^{an} - s_i^{ap} + \alpha } \right]}_ + }}
\end{equation}
where the operator ${\left[ x \right]_ + } = \max (x,0)$. It is crucial to select ``hard'' triplets that do not fulfill the constraint in Equation (\ref{equ:st.}).

Training examples are organized as anchor-positive (AP) pairs of same-speaker feature chunks.
Mini-batches are formed by picking $N$ pairs and splitting them onto $M$ GPUs, so that each GPU has $N/M$ pairs.
All AP pairs in the mini-batch are used, and anchor negatives are selected from the same batch, though not necessarily the same minibatch.
Why do we search across GPUs for negative examples?
In the beginning of training, it may be easy to find hard negatives which cannot fulfill the constraint in Equation (\ref{equ:st.}).
However, finding hard negatives becomes harder and harder as training progresses.
Thus, we search over the entire batch for negative examples, rather than in the same minibatch.

\begin{table}[t]
\caption{\it The effect of the number of minibatches scanned to pick a negative
sample, measured by the probability of finding a hard sample (Equation \ref{equ:st.} not met) and training time differential.}
%\footnotesize
\label{tab:selectN}
%\vspace{-2mm}
\begin{center}
\begin{tabular}{c|c|c|c|c}
\hline
\#GPUs & 1 & 4 & 8 & 16 \\
\hline
Prob(hard) & 29.06\% & 43.29\% & 45.54\% & 50.99\% \\
\hline  %\vspace{1pt}
Rel. time cost& - & +5.47\% & +6.09\% & +15.28\% \\
\hline  %\vspace{1pt}
\end{tabular}%
\end{center}
\end{table}

We design a simple experiment to investigate the effects of the number of GPUs scanned when picking negative
samples. We trained the model for 6 epochs (about 1/4th of total training
time), with $N/M=64$ utterances per mini-batch. We see that the probability of finding an hard negative sample rises without too much additional time cost as the number of GPUs scanned increases (see
Table \ref{tab:selectN}).
For example, when the number of GPUs increases from 1 to 4, the probability of finding an effective negative sample increases by 48.97\%,
while time costs only increase by 5.47\%.

\subsection{Softmax Pre-training}
\label{sec:pretrain}

To avoid suboptimal local minima early-on in training, \cite{facenet} proposed using
\emph{semi-hard} negative exemplars, as they are further away from the anchor than the positive exemplar,
but still hard because the AN cosine similarity is close to the AP cosine similarity. That is to say  $s_i^{an} + \alpha > s_i^{ap} > s_i^{an}$. In the beginning epochs, the model trains only using \emph{semi-hard}
negative exemplars, followed by the normal training with hard negative ones. However, properly scheduling ``semi-hard'' samples is not simple because of model and dataset variability.
Instead of the \emph{semi-hard} train stage, we use a softmax and cross entropy loss to
pre-train the model.
It uses a classification layer \cite{softmax} to replace the length normalization and triplet loss layers in the standard Deep Speaker
architecture described in Figure \ref{fig:archit}.

Using softmax pre-training to initialize the weights of the network has two main benefits.
First, we notice that the cross entropy loss produces stabler convergence than triplet loss.
We hypothesize that this is because softmax training is not impacted by the variable difficulty of pairs in triplet loss.
Secondly, while triplet selection is faster with larger minibatches, smaller mini-batches usually yield better generalization in Stochastic Gradient Descent (SGD) \cite{sgd}.

\begin{figure}[t]
\centering
    \begin{subfigure}{\linewidth}
    %\begin{minipage}{8cm}
    \centering
    
    \includegraphics[width=8cm]{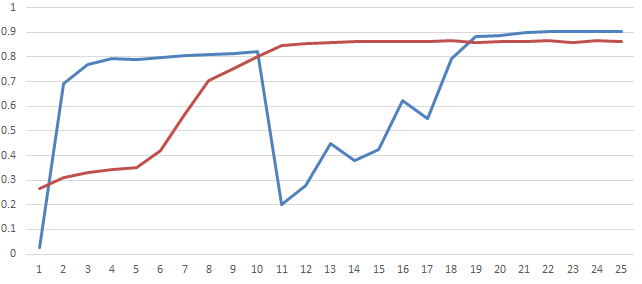}
    %\end{minipage}
    \caption{ACC}
    \label{fig:pretrain-aac}
    \end{subfigure}

    %\subfigure[EER]{
    \begin{subfigure}{\linewidth}
    %\begin{minipage}{8cm}
    \centering
    \includegraphics[width=8cm]{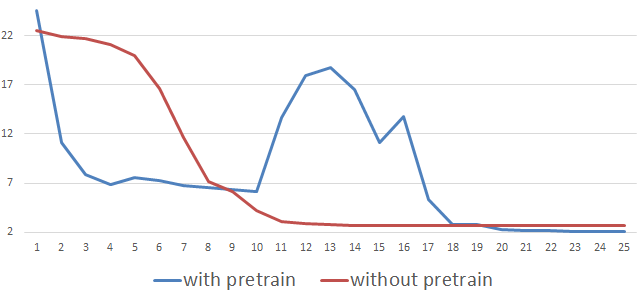}
    %\end{minipage}
    \caption{EER}
    \label{fig:pretrain-eer}
    \end{subfigure}
    %}

%\vspace{-20pt}
\caption{\it The effect of softmax pre-training, trained and measured in Train50k and Eva200 dataset,
 using ResCNN network
 architecture. (a) accuracy (ACC) vs epoch trained and (b) equal error rate (EER) vs epoch trained}
\label{fig:pretrain}
\end{figure}

We designed an experiment to investigate the effect of softmax pre-training for Deep Speaker. Figure \ref{fig:pretrain} shows the validation set accuracy ratio (ACC) and equal error rate (EER) vs epoch index (details about experimental setup are described in Section
\ref{sec:experimental_setup}).
We pre-train using softmax for 10 epochs, followed by 15 epochs triplet loss
training, which causes the spikes at epoch 11 in Figure
\ref{fig:pretrain-aac} and Figure \ref{fig:pretrain-eer}.
With the whole 25 epochs training, the pre-trained neural network can achieve lower EER and higher ACC than neural networks without pre-training. More detailed comparison will be described
in Section \ref{sec:experiments}).

\section{Experimental Setup}
\label{sec:experimental_setup}

\subsection{Dataset and Evaluation}
\label{sec:dataset}
To investigate the performance of Deep Speaker, we run both speaker verification and identification experiments on three
different datasets: UIDs (including Train250k, Train50k and Eva200), XiaoDu, and MTurk.
The first two datasets consist of Mandarin speech recorded by mobile phone apps, while the third is English speech collected on Amazon Mechanical Turk. 
UIDs and MTurk are text-independent datasets, while XiaoDu 
 is text-dependent. The details of the three datasets are given as follows. 
%Speech is sampled at 16kHz and compressed using same internal process.

\begin{table}[t]
\caption{Statistics of datasets used in the experiments. {\it UIDs} and {\it XiaoDu} consist of Mandarin speech from mobile queries, while {\it MTurk} is a English dataset from Amazon Mechanical Turk. {\it UIDs} and {\it Mturk} are text independent, while {\it XiaoDu} is a text-dependent dataset consisting of wake-words. }
\begin{center}

\begin{tabular}{c|cccc}
\hline
 & \#spkr & \#utt & \#utt/spkr & dur/utt \\
\hline
Train250k & 249,312 & 12,202,181 & 48.94 &  3.72s \\
Train50k & 46,835 & 2,236,379 & 47.75 &  3.66s \\
\hline  %\vspace{1pt}
Eva200 & 200 & 3,800 & 19 & 4.25s \\
\hline  %\vspace{1pt}
\end{tabular}%
\caption*{\it UIDs dataset.}
\label{tab:cuid}

\begin{tabular}{c|cccc}
\hline
 & \#spkr & \#utt & \#utt/spkr & dur/utt \\
\hline
train & 11,558  & 89,227  & 7.72  & 1.56s  \\
\hline  %\vspace{1pt}
test & 844 & 10,014 & 11.86 & 1.48s \\
\hline  %\vspace{1pt}
\end{tabular}%
\caption*{\it XiaoDu dataset}.
\label{tab:XiaoDu}

\begin{tabular}{c|cccc}
\hline
 & \#spkr & \#utt & \#utt/spkr & dur/utt \\
\hline
train &  2,174  & 543,840  & 250.16 & 4.16s  \\
\hline  %\vspace{1pt}
test & 200 & 4,000  & 20 & 4.31s  \\
\hline  %\vspace{1pt}
\end{tabular}%
\caption*{\it MTurk dataset.}
%\footnotesize
\label{tab:english}

\end{center}
\end{table}

\begin{itemize}[leftmargin=.2in]
    \item
    {\it UIDs} is collected from anonymized voice search queries. Table \ref{tab:cuid} lists statistics for the utterances, the durations of which mostly range from 3 to 5 seconds. The full training partition, {\it Train250k}, comprises 249,312 speakers, 12,202,181 utterances, and more than 12,600 hours of speech. The subset {\it Train50k} comprises 46,835 speakers, 2,236,379 utterances, and more than 2,270 hours of speech. The evaluation partition, {\it Eva200}, consists of 200 speakers that do not overlap with the training speakers. 380,000 speaker recognition trials were run.
    \item
    {\it XiaoDu} contains Baidu wake-word utterances, ``Xiaodu, xiaodu". The full training dataset comprises 11,558 speakers, 89,227 utterances, and more than 38 hours of speech. The evaluation dataset consists of 844 speakers that do not overlap with the training speakers. 1,001,400 speaker recognition trials were run.
    \item
    {\it MTurk} contains scripted English utterances collected on Amazon Mechanical Turk. The median utterance length is around 4 seconds, and the 25th and 75th percentile lengths are $\sim$3 seconds and $\sim$5 seconds. The full training dataset comprises  2,174 speakers, 543,840 utterances,  and more than 620 hours of speech. The evaluation dataset consists of 200 speakers that do not overlap with the training speakers. 400,000 speaker recognition trials were run.
\end{itemize}

Speaker verification and identification trials were constructed by randomly picking one anchor positive sample (AP) and 99 anchor negative samples (AN) for each anchor utterance. 
Then, we computed the cosine similarity between the anchor
sample and each of the non-anchor samples. EER and ACC are used for speaker verification and identification, respectively.
Since gender labels are not available in all our datasets, the evaluation here is not split by gender, unlike the NIST speaker recognition evaluations (SREs) in \cite{ivector}.

%\section{Experiments}

\subsection{Baseline DNN i-vector System}
\label{sec:baseline}

The baseline DNN i-vector model is built based on \cite{dnn-ivector}.
Raw audio is converted to 40-dimensional log mel-filter bank (Fbank) coefficients and 20-dimensional MFCC with a frame-length of 25ms.
Delta and acceleration are appended to the input, and a frame-level energy-based VAD selects features corresponding to speech frames.

A seven-layer DNN that contains 600 input nodes, 1024 nodes in each hidden layer, and 4,682 output nodes is
trained with cross entropy using the alignments from a HMM-GMM model. The input layer of the DNN is
composed of 15 frames (7 frames on each side of the frame for which predictions are made)
where each frame corresponds to 40 dimension Fbank coefficients. The DNN is used to provide
the posterior probability in the proposed framework for the 4,682 senones defined by a decision tree.

A 1024 diagonal component UBM is trained in a gender-independent fashion, along with a 400 dimensional
i-vector extractor, followed by length normalization and PLDA.

\subsection{Training Methodology}
\label{sec:training}

Deep Speaker models are trained using the SpeechDL \cite{ds2} distributed machine learning system with 16 K40 GPUs.
Audio is converted to 64-dimensional Fbank coefficients, normalized to have zero mean and unit variance.
The same VAD processing as DNN i-vector baseline system is used here.

As described in Section \ref{sec:pretrain}, Deep Speaker models are trained in two stages: softmax pre-training and triplet
loss fine-tuning. In both stages, we use synchronous SGD with 0.99 momentum \cite{sgd}, with a linear decreasing learning rate schedule from 0.05 to 0.005. The model is pre-trained for 10 epochs with a minibatch size of 64 and fine-tuned with triplet loss for 15 epochs using a minibatch size of 128.
Training pairs are re-shuffled in each epoch. The margin $\alpha$ is set to 0.1 in \ref{equ:st.}
A development dataset is used for hyper-parameter tuning and early stopping.

\section{Experimental Results}
\label{sec:experiments}

\subsection{Speaker-independent Experiments on UIDs}
\label{sec:speaker independent experiment}

First, we compare the DNN-based i-vector system and Deep Speaker on the UIDs dataset, with the Train50k partition as the training set and Eva200 partition as the test set. Deep Speaker models are trained with different neural network architectures (GRU or ResCNN), and different strategies (softmax, triplet loss, or softmax + triplet loss, which is softmax pretraining followed by triplet loss fine-tuning). The results are listed in Table \ref{tab:exp Train50k}. All Deep Speaker models achieve notable improvements over the baseline, roughly 50 - 80\%
relative reduction on EER, and 60 - 70\% relative improvement on ACC.

\subsubsection{Softmax Pre-training }

Training using ``softmax + triplet" loss achieves the best
performance, followed by triplet loss-only and softmax-only training, in decreasing performance order.
In particular, the ResCNN with softmax + triplet loss achieves
63.62\% and 17.10\% relative reduction on EER and 47.53\% and 31.33\% on error (1$-$ACC) compared to softmax-only and triplet loss-only models.
The pre-trained GRU architecture achieves 48.89\% and 14.24\% relative reduction on EER and 38.05\% and 33.59\% on identification error compared to the other two GRU systems. The results confirm the advantages of end-to-end training and softmax pre-training.

\begin{table}[t]
\caption{\it The performance of Mandarin text-independent speaker recognition task using Train50k as train
set and Eva200 as test set. Results from different neural network architectures and training methodologies are reported.}
%\footnotesize
\label{tab:exp Train50k}
%\vspace{-2mm}
\begin{center}
\begin{tabular}{l|c|c}
\hline
 system & EER[\%] & ACC[\%] \\
\hline
DNN i-vector baseline &  13.79  &  51.72    \\
\hline  %\vspace{1pt}
ResCNN, softmax & 6.13 & 81.95 \\
ResCNN, triplet & 2.69 & 86.21 \\
ResCNN, softmax (pre-train) + triplet & \textbf{2.23} & \textbf{90.53} \\
\hline  %\vspace{1pt}
GRU, softmax & 5.42 & 83.05   \\
GRU, triplet & 3.23 & 84.19  \\
GRU, softmax (pre-train) + triplet & \textbf{2.77} &  \textbf{89.50}   \\
\hline  %\vspace{1pt}
\end{tabular}%
\end{center}
\end{table}

\subsubsection{Network Architecture}

While the GRU architecture outperforms ResCNN with softmax-only training, ResCNN outperforms GRU layers after triplet loss fine-tuning.
As shown in Table \ref{tab:exp Train50k},
GRU achieves a 11.58\% lower EER and a 6.09\% lower error rate compared to ResCNN after softmax training.  After triplet loss training, ResCNNs had a 19.49\% lower EER and a 10.88\% lower error rate than GRUs.  
As ``softmax + triplet loss" training strategy achieves best performance for both GRU and ResCNN, we will use it in the following experiments and also omit this label for brevity's sake.

Time cost is another important consideration for choosing network architectures.
We measure the training speed of a model as the number of minibatches that can be processed per second.
In this experiment, ResCNN can processes 0.23 minibatches per second, while GRU processes 0.44 minibatches per second.
The time cost gap could be partially caused by using Deep Speech 2 HPC techniques for the GRU network \cite{ds2} and not spending comparable effort speeding up the ResCNN.

\subsubsection{System Combination}

Individual Deep Speaker models perform well separately, but we anticipate that the ResCNN and GRU systems can benefit from fusion because of their contrasting architectures. To fuse
ResCNN and GRU, we investigate two methods: embedding fusion and score fusion. In the first method,
we add the speaker embedding from both models together, followed by length normalization and cosine
score. In the second method, we first normalize the scores using mean and variance calculated from all
scores and add them together. Table \ref{tab:exp fusion} indicates that relative to the best single system (ResCNN),  both fused system
improve the single-system baselines. Especially, the score fusion method gets the best performance, with 7.17\% and 13.37\% reductions in EER and error, respectively.

\begin{table}[t]
\caption{\it The system fusion performance of text-independent speaker recognition task using
Train50k as train set, Eva200 as test set.}
%\footnotesize
\label{tab:exp fusion}
%\vspace{-2mm}
\begin{center}
\begin{tabular}{l|c|c}
\hline
 system & EER[\%] & ACC[\%] \\
\hline  %\vspace{1pt}
ResCNN  & 2.23 & 90.53 \\
GRU   & 2.77 &  89.50   \\
\hline  %\vspace{1pt}
embedding fusion & 2.17 &  90.95   \\
score fusion & \textbf{2.07} & \textbf{91.83} \\
\hline  %\vspace{1pt}

\end{tabular}%

\end{center}
\end{table}

\subsubsection{Amount of Training Data }

\begin{table}[t]
\caption{\it The performance of text-independent speaker recognition task using both Train250k and Train50k as training
sets and Eva200 as the test set.
All deep speaker models
are trained with softmax pre-training.}
%\footnotesize
\label{tab:exp Train250k}
%\vspace{-2mm}
\begin{center}
\begin{tabular}{l|c|c}
\hline
 system & EER[\%] & ACC[\%] \\
\hline  %\vspace{1pt}
ResCNN on Train50k & 2.23 & 90.53 \\
ResCNN on Train250k & \textbf{1.83} & \textbf{92.58} \\
\hline  %\vspace{1pt}
GRU  on Train50k & 2.77 &  89.50    \\
GRU  on Train250k & \textbf{2.35} & \textbf{90.77} \\
\hline  %\vspace{1pt}

\end{tabular}

\end{center}
\end{table}

Table \ref{tab:exp Train250k} shows the impact of training dataset size on speaker recognition performance.
We do not experiment with the i-vector baseline here, as it's too time consuming and computationally expensive. 
\footnote{For example, the total variance matrix $T$ in the i-vector model is too hard to compute on a big dataset. In practice, people usually train the i-vectors using subsets of a large dataset. Indeed, we tried training i-vector systems on larger datasets and got no obvious improvements.}
It is clear that using tens of millions of samples results in a performance boost. Compared to using only around 1/5th the data, the using the full dataset reduces the identification error and EER by 17.94\% and 21.65\% for ResCNN and 15.16\% and 13.88\% for GRU.

\subsubsection{Utterances Number for Enrollment }

\begin{table}[t]
\caption{\it Effect of enrollment utterance count.  The performance column is formatted as
EER/ACC in each cell. All models are trained on Train50k dataset. }
%\footnotesize
\label{tab:exp enroll}
%\vspace{-2mm}
\begin{center}
\begin{tabular}{l|c|c|c}
\hline
 \#utt & i-vector & ResCNN & GRU \\
 \hline
1 &  13.79 / 51.72 & 2.23 / 90.53 & 2.77 / 89.50 \\
2 &  10.37 / 63.21 & 1.39 / 95.36 & 1.70 / 94.64  \\
3 &   8.21 / 71.04 & 1.29 / 96.56 & 1.56 / 96.47  \\
5 & \textbf{7.57 / 75.02} &  \textbf{1.13 / 96.83} &  \textbf{1.37 / 97.07}  \\
\hline  %\vspace{1pt}
\end{tabular}%

\end{center}
\end{table}

To investigate the effect of the enrollment utterance  count on recognition tasks, we choose 1 to 5 utterances for each person's enrollment. 
The speaker embeddings are produced by averaging the enrollment utterance embeddings. As before, speaker verification and identification trials were constructed by randomly
picking one AP and 99 AN speaker embeddings for each anchor utterance. In
total, 280,000 trials were conducted.

Table \ref{tab:exp enroll} shows that the EER decreases and ACC increases as the enrollment utterance count increases, though with diminishing returns.  These results have implications for production speech recognition system design choices.  In particular, using too many enrollment utterances would provide minimal performance gains while increasing inference time, making new user enrollment more cumbersome, and increasing memory usage.

\subsection{Text-dependent Experiments on XiaoDu}
\label{sec:text dependent experiment}
% xiaoduxiaodu

Table \ref{tab:exp xiaodu} shows the performance of both ResCNN and GRU models for text-dependent speaker recognition on the XiaoDu dataset. The flag ``on Train50k'' means the  Deep Speaker models are only trained on Train50k, while the flag ``finetuned'' indicates that we first trained the model on Train50k, then used the XiaoDu dataset to fine-tune with triplet loss for Deep Speaker systems, and i-vector extraction for the DNN i-vector system.

 \begin{table}[t]
\caption{\it The text-dependent speaker recognition performance of different systems on
XiaoDu dataset.}
%\footnotesize
\label{tab:exp xiaodu}
%\vspace{-2mm}
\begin{center}
\begin{tabular}{l|c|c}
\hline
 system & EER[\%] & ACC[\%] \\
\hline

DNN i-vector & \textbf{3.50}   & \textbf{95.05}    \\
ResCNN  & 4.10 & 93.08 \\
GRU  & 3.82 & 93.75   \\
\hline  %\vspace{1pt}
ResCNN on Train50k & 3.62 & 93.25  \\
GRU on Train50k    & 3.74 & 94.45  \\
\hline  %\vspace{1pt}
finetuned DNN i-vector &  3.40   & 94.75     \\
finetuned ResCNN & 2.83  &  94.85 \\
finetuned GRU & \textbf{2.78} &  \textbf{95.75}   \\
\hline  %\vspace{1pt}

\end{tabular}%
\end{center}
\end{table}

Interestingly, the DNN i-vector baseline system achieves the best performance when only using XiaoDu
to train the models. There are two possible reasons here. Firstly, the XiaoDu dataset is too small to train complex deep models like Deep Speaker, and we may be overfitting. Secondly, the text-dependent speaker verification task constrains the lexicon and phonetic variability, so the i-vector extractor based on factor analysis can cover the speaker variability in the small dataset better.

To our surprise, we find that Deep Speaker models only trained on Train50k achieve slightly better performance than models only trained on XiaoDu. That is to say, models trained using text-independent datasets can work well in text-dependent tasks. We believe that the superior performance is a result of the amount of training data.   

Fine-tuning the traditional DNN i-vector system does not significantly improve performance, while fine-tuned ResCNN and GRU networks outperform the DNN i-vector system, by 16.76\% and 18.24\% relative reduction on EER, and similar ACC. 
This shows that pre-training on large text-independent datasets can aid in data-constrained text-dependent scenarios.
We speculate that the large datasets can cover a greater diversity of samples and encourage model generalization.

\subsection{Text-independent Experiments on MTurk }
\label{sec:english experiment}

The MTurk experimental results in Table \ref{tab:exp english} showcase that Deep Speaker works across languages.
The flag ``on Train50k'' means the  Deep Speaker models are only trained on Train50k, while the flag ``finetuned'' means models are first trained on Train50k, and then fine-tuned on the MTurk dataset using triplet loss. Note that this is a non-trivial task, since Mandarin and English speech sounds disparate.
We don't report a fine-tuned result for DNN i-vector baseline here.  Because Mandarin and English have different phone sets, the ASR-DNN model is difficult to adapt.

 \begin{table}[t]
\caption{\it The text-independent speaker recognition performance of different systems on
MTurk dataset.}
%\footnotesize
\label{tab:exp english}
%\vspace{-2mm}
\begin{center}
\begin{tabular}{l|c|c}
\hline
 system & EER[\%] & ACC[\%] \\
\hline
DNN i-vector &  3.88 & 89.68      \\
ResCNN  & \textbf{3.41} & 91.23  \\
GRU  & 3.50  & \textbf{91.68}  \\
\hline  %\vspace{1pt}
ResCNN on Train50k & 5.92 & 85.41  \\
GRU on Train50k    & 5.57 & 88.22  \\
\hline  %\vspace{1pt}
finetuned ResCNN & 2.68 & 94.53  \\
finetuned GRU & \textbf{2.40} &  \textbf{94.88}   \\
\hline  %\vspace{1pt}

\end{tabular}%
\end{center}
\end{table}

By comparing the different systems trained only on MTurk, the ResCNN and GRU system reduce EER by
12.11\% and 9.79\% and error by 15.02\% and 19.38\% compared to the DNN i-vector system.
Interestingly, models trained solely on the Mandarin Train50k dataset perform fairly well on English speaker classification, even without fine-tuning.
The ``finetuned'' models outperform ``non-finetuned'' models, reducing the EER by 25\% and the error rate by 35\%.

These results indicate that Deep Speaker systems can work well not only on Mandarin, but also across other languages. In addition, the representations learned by Deep Speaker transfer well across different languages.

\subsection{Time Span Experiments on UIDs}
\label{sec:the shift experiment}

Speaker recognition systems usually struggle with robustness to time between enrollment and test time.
People's voiceprints change over time, just like their appearances. 
We test the robustness of our model across a wide range of time spans using the Eva200 dataset. 
In Table \ref{tab:exp timeshift}, the first column divides the different time spans, ``1 week'' means the time span of registration and verification is less than 1 week, ``1 month'' means less than 1 month but longer than 1 week, and ``3 months'' means  less than 3 months but longer than 1 month.

The performance of all systems decrease as the time span between enrollment and test increases, but ResCNN can still achieve the best performance with the same time span.

\begin{table}[t]
\caption{\it Effect of time span on recognition performance across models. The performance column 
is formatted with EER/ACC in each cell. All models are trained on Train50k dataset }
%\footnotesize
\label{tab:exp timeshift}
%\vspace{-2mm}
\begin{center}
\begin{tabular}{l|c|c|c}
\hline
 \#time span & baseline & ResCNN & GRU \\
 \hline
1 week & 12.73 / 55.33   & \textbf{2.11 / 91.11} & 2.63 / 90.66  \\
1 month & 14.39 / 46.96  & \textbf{2.50 / 88.66} & 3.33 / 87.57  \\
3 months & 15.31 / 44.37 & \textbf{2.76 / 87.45} & 3.42 / 85.80  \\
\hline  %\vspace{1pt}
\end{tabular}%

\end{center}
\end{table}

\section{Conclusion}
\label{sec:conclusion}
In this paper we present a novel end-to-end speaker embedding scheme, called Deep Speaker. The proposed system  directly learns a mapping from speaker utterances to a  hypersphere where
cosine similarities directly correspond to a measure of speaker similarity. We experiment with two different
neural network architectures (ResCNN and GRU) to extract the frame-level acoustic features. A
triplet loss layer based on cosine similarities is proposed for metric learning, along with a
batch-global negative selection across GPUs. Softmax pre-training is used for achieving better performance.

The experiments show that the Deep Speaker algorithm significantly improves  the text-independent
speaker recognition system as compared to the traditional DNN-based i-vector approach.
In the Mandarin dataset UIDs, the EER decreases roughly 50\% relatively, and error
decreases by 60\%. In the English dataset MTurk, the equal error rate decreases by 30\%
relatively, and error decreases by 50\%. Another strength of Deep Speaker is that it can take full advantage of transfer learning
to solve the speaker recognition problems on small data sets, for both text-independent and text-dependent
tasks.

Future work will focus on better understanding the
error cases, reducing model size, and reducing CPU requirements.
We will also look into ways of improving the long training times.

\section{Acknowledgments}
We would like to thank Liang Gao and Yuanqing Lin for their supports and great insights on speaker recognition.
We would also like to thank Sanjeev Satheesh, Adam Coates, and Andrew Ng for useful discussions and thoughts.
Also our work would not have been possible without the data support of Hongyun Zeng, Yue Pan, Jingwen Cao, and Limei Han.

%\vspace{10pt}

\end{sloppy}

\end{document}